\RequirePackage{fix-cm}

\RequirePackage{snapshot}

\documentclass[twocolumn]{svjour3}          
\smartqed  

\newcommand{\minisection}[1]{\vspace{0.04in} \noindent {\bf #1}\ \ }
\usepackage{subcaption}
\usepackage{multirow}
\usepackage{multicol}

\usepackage{amsmath,amsfonts,bm}

\def\eqref#1{equation~\ref{#1}}

\def\1{\bm{1}}

\DeclareMathAlphabet{\mathsfit}{\encodingdefault}{\sfdefault}{m}{sl}
\SetMathAlphabet{\mathsfit}{bold}{\encodingdefault}{\sfdefault}{bx}{n}

\usepackage{graphicx}%
\usepackage{amsmath,amssymb,amsfonts}%
\usepackage{amsthm}%
\usepackage{mathrsfs}%
\usepackage[title]{appendix}%
\usepackage{xcolor}%
\usepackage{colortbl}
\usepackage{textcomp}%
\usepackage{manyfoot}%
\usepackage{booktabs}%
\usepackage{algorithm}%
\usepackage{algorithmicx}%
\usepackage{algpseudocode}%
\usepackage{listings}%
\usepackage{url}
\usepackage{capt-of}
\usepackage{multirow}
\usepackage{makecell}
\usepackage{wrapfig}
\usepackage{array}
\usepackage{amssymb}
\usepackage{pifont}%
\usepackage{soul, xcolor}
\setstcolor{red}

\usepackage{array}
\makeatletter
\newcommand*{\@rowstyle}{}
\newcommand*{\rowstyle}[1]{
  \gdef\@rowstyle{#1}%
  \@rowstyle\ignorespaces%
}

\newcolumntype{=}{
  >{\gdef\@rowstyle{}}%
}

\newcolumntype{+}{
  >{\@rowstyle}%
}

\makeatother

\usepackage[round]{natbib}
\usepackage[final, colorlinks=true,allcolors=blue,breaklinks=true]{hyperref} 

\definecolor{f_trail}{RGB}{170,170,170}
\definecolor{f_grass}{RGB}{0,255,0}
\definecolor{f_vegetation}{RGB}{102,102,51}
\definecolor{f_sky}{RGB}{0,120,255}
\definecolor{f_obstacle}{RGB}{0,0,0}
\definecolor{u_Bed}{RGB}{0,0,255}
\definecolor{u_Books}{RGB}{233, 89, 48}
\definecolor{u_Ceiling}{RGB}{0, 218, 0}
\definecolor{u_Chair}{RGB}{149, 0, 240}
\definecolor{u_Floor}{RGB}{222, 241, 24}
\definecolor{u_Furniture}{RGB}{255, 206, 206}
\definecolor{u_object}{RGB}{0, 224, 229}
\definecolor{u_Picture}{RGB}{106, 136, 204}
\definecolor{u_Sofa}{RGB}{117, 29, 41}
\definecolor{u_Table}{RGB}{240, 35, 235}
\definecolor{u_Tv}{RGB}{0, 167, 156}
\definecolor{u_wall}{RGB}{250, 139, 0}
\definecolor{u_Window}{RGB}{225, 229, 195}

\begin{document}
\sloppy
\title{Adapter-Enhanced Semantic Prompting for Continual Learning}

\author{Baocai Yin, Ji Zhao, Huajie Jiang, Ningning Hou, Yongli Hu, \\
        Amin Beheshti, Ming-Hsuan Yang, Yuankai Qi
}
\titlerunning{Adapter-Enhanced Semantic Prompting for Continual Learning}
\authorrunning{Cotogni et al.} 

\institute{Baocai Yin, Ji Zhao, Huajie Jiang and Yongli Hu are with Beijing Key Laboratory of Multimedia and Intelligent Software Technology, Faculty of Information Technology, Beijing University of Technology, Beijing, China. \email{ybc@bjut.edu.cn, zhaoji@emails.bjut.edu.cn, jianghj@bjut.edu.cn, huyongli@bjut.edu.cn}. \\
Ming-Hsuan Yang is with the Department of Electrical Engineering and Computer Science, University of California at Merced, Merced, USA. \\
Ningning Hou, Amin Beheshti and Yuankai Qi are with the School of Computing, Macquarie  University, Sydney, NSW, Australia. \\
(Corresponding author: Huajie Jiang, Yuankai Qi)
}


\maketitle

\begin{abstract}
  Continual learning (CL) is essential for enabling models to adapt to dynamic data streams, with the primary challenge being the mitigation of catastrophic forgetting of previously acquired knowledge. Recent advancements have highlighted the potential of prompt-based CL methodologies, which have shown promise in preserving the knowledge embedded within pre-trained models, thereby addressing catastrophic forgetting. However, these methodologies are constrained by their focus on learning visual prompts for individual tasks, resulting in suboptimal generalization to unseen categories.
  
  To enhance the model's ability to retain knowledge from prior tasks while simultaneously improving generalization to new tasks, we introduce a novel, lightweight framework for CL termed Adapter-Enhanced Semantic Prompting (AESP). This framework leverages class semantic information to augment visual features, thereby establishing relational links among different categories. This not only fortifies the retention of prior knowledge but also facilitates adaptation to new tasks. To seamlessly integrate visual and semantic information, we have developed a lightweight Adapter-enhanced Vision Transformer (ViT) architecture specifically designed for feature adaptation. Furthermore, we have implemented a robust Query-Key matching strategy to select the optimal task prompt pair, thereby enhancing the accuracy of final predictions. Comprehensive experiments conducted across three benchmark continual learning datasets demonstrate that our proposed framework outperforms several state-of-the-art approaches, showcasing its efficacy and robustness.
\end{abstract}


\section{Introduction}\label{sec1}
Continual learning (CL) ~\citep{aljundi2018memory,belouadah2021comprehensive,de2021continual, tian2023survey,DBLP:journals/ijcv/ZhouCYZL25,wang2024classsurvey} is a paradigm designed to empower neural network models to seamlessly learn from evolving data streams and adapt to new environments. Unlike traditional machine learning models ~\citep{jordan2015machine, mahesh2020machine}, which are trained on static datasets and remain unchanged post-training, CL facilitates the accumulation of existing knowledge while assimilating new information from successive tasks. This dynamic approach obviates the necessity of retaining all historical task data and retraining models from scratch ~\citep{de2021continual, gao2024beyond}, thereby addressing pivotal challenges such as data privacy and memory resource constraints. The inherent advantages of CL have spurred significant research interest in recent years ~\citep{lopez2017gradient,parisi2019continual,ren2019incremental}.

Despite the advantage of CL, it is beset by the phenomenon of catastrophic forgetting ~\citep{mccloskey1989catastrophic,french1999catastrophic,DBLP:journals/ijcv/KongLQWT23}, wherein the acquisition of new knowledge results in the overwrite of prior knowledge, leading to substantial performance degradation on previously learned tasks. To mitigate this issue, several strategies ~\citep{hou2019learning,tao2020topology} have been proposed, including the retention of representative samples from past tasks for knowledge replay. Upon the arrival of a new task, these retained samples are rehearsed in conjunction with the new data, effectively reducing the model’s propensity to forget. Nevertheless, this method introduces a memory burden and potential privacy risks. Alternatively, approaches such as ~\citep{de2021continual,bojian2023continual} advocate for the addition of new branches to the network for each incoming task. While this strategy aids in knowledge preservation, it engenders an increase in model size, thereby compromising inference speed and overall efficiency.

    \begin{figure*}[t]
		\centering
		\includegraphics[width=\textwidth]{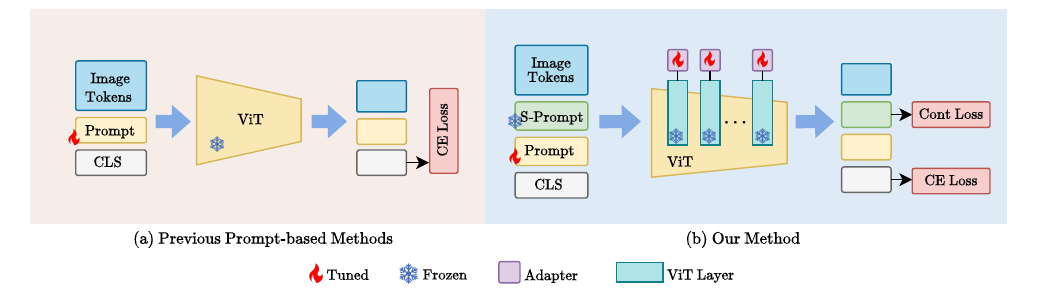}
        \vspace{-6mm}
        \caption{Comparison between previous prompt-based approaches and our framework. Previous approaches mainly use visual prompts to update the image features. In contrast, our method introduces semantic prompts (S-Prompts) to enrich the semantic information and embeds a fine-tunable adapter structure to effectively learn adaptive image features.}
		\label{fig:motivation}
        \vspace{-2mm}
	\end{figure*}

In recent years, prompt-based methodologies ~\citep{wang2022learning,wang2022dualprompt, tang2022learning,khan2023introducing,DBLP:journals/ijcv/LiuPZ24,} have emerged as promising approaches for addressing the challenges of continual learning (CL). As illustrated in Figure ~\ref{fig:motivation} (a), these techniques integrate learnable visual prompts into image tokens during the training phase. The primary objective is to adapt pre-trained Vision Transformer (ViT) models to acquire new knowledge across successive CL tasks, while simultaneously mitigating forgetting by maintaining the ViT parameters in a frozen state. Distinct from traditional methods, prompt-based approaches necessitate the training of fewer parameters, yet they achieve superior performance by harnessing the extensive pre-trained knowledge embedded within ViT models. 
    
Nevertheless, a significant limitation of these methods is their reliance on visual information alone, which can constrain the model’s capacity to generalize effectively, especially when operating with limited and diverse training data. This constraint often results in ambiguity between newly acquired and previously learned categories, thereby exacerbating issues related to catastrophic forgetting. To address this limitation, leveraging class semantic information derived from large language models (LLMs) offers a more generalized and adaptable framework across diverse tasks ~\citep{khan2023introducing,park2024pre}. Notably, models such as CLIP ~\citep{radford2021clip} facilitate the alignment of samples within the same class with their corresponding semantic information, thereby constructing a more generalized feature space. This alignment significantly enhances the model’s robustness when encountering unseen samples from the same class, as it encourages the model to focus on shared semantic features rather than relying solely on specific visual cues.

In this work, we introduce a novel Adapter-Enhanced Semantic Prompting (AESP) framework, specifically designed to advance the field of continual learning (CL). Unlike prior methods that focus solely on learning visual prompts, our approach leverages semantic embeddings derived from BERT’s text encoder ~\citep{kenton2019bert} as semantic prompts, thereby enhancing the generalization of features encoded by the pre-trained Vision Transformer (ViT) model. However, a significant discrepancy persists between the representation spaces of semantic and visual information. The inherent attention mechanisms within a pre-trained ViT may not adequately reconcile these distinct modalities. To address this challenge, we introduce compact, trainable adapter modules within the ViT architecture, which are engineered to effectively fuse visual and semantic information. As illustrated in Figure ~\ref{fig:motivation} (b), these adapters are integrated into each layer of the ViT, enabling the model to better accommodate semantic information and fostering a more robust interaction between visual and semantic feature spaces. The AESP framework not only augments the integration of multimodal features but also endows the learned knowledge with enhanced generalization capabilities, thereby bolstering the model’s adaptability to subsequent tasks. Moreover, to ensure the precise selection of task-specific prompts for feature adaptation, we develop a novel Prompt-Key Matching (PKM) method. This method synthesizes a triad of matching strategies, ensuring the selection of the most relevant prompts from the repository, thereby optimizing task-matching accuracy and efficiency. Extensive experimental evaluations across various incremental tasks demonstrate the efficacy of our proposed AESP framework in significantly improving model performance.

The main contributions of our work are summarized as follows:
\begin{itemize}
    \item We propose the Adapter-Enhanced Semantic Prompting (AESP) framework, a novel approach for continual learning that integrates multi-modality prompts and adapters to facilitate adaptive feature learning and mitigate catastrophic forgetting.
    \item We innovate by utilizing semantic prompts to enhance the generalization of visual features, and we design a semantic contrast loss to ensure semantic consistency, thereby improving the model's knowledge transfer and accumulation capabilities.
    \item We develop a sophisticated Prompt-Key Matching (PKM) mechanism that employs multiple strategies to enhance the accuracy and efficiency of task-specific prompt selection.
    \item We conduct comprehensive experiments on three benchmark continual learning datasets, demonstrating that our AESP framework outperforms several state-of-the-art (SOTA) approaches, thereby establishing its superior performance and robustness.
\end{itemize}

\section{Related Work}

\minisection{Continual Learning.} Continual Learning (CL) is a pivotal subdomain within the realm of machine learning, focused on the development of models capable of sequentially acquiring new knowledge while preserving previously acquired information ~\citep{belouadah2021comprehensive, de2021continual, DBLP:journals/ijcv/CotogniYCBW25}. The core challenge in this domain is the propensity for the acquisition of new knowledge to interfere with prior learning, resulting in the phenomenon known as catastrophic forgetting ~\citep{mccloskey1989catastrophic, shi2021overcoming}. To mitigate this issue, CL techniques have evolved to incorporate a variety of strategies aimed at balancing the retention of existing knowledge with the assimilation of new information. 
Regularization methods represent one such strategy, wherein critical parameters are preserved by imposing constraints, thereby forestalling substantial modifications that could overwrite prior learning ~\citep{nusrat2018comparison, ahn2019uncertainty, zhang2020regularize}. Knowledge distillation is another prominent approach, frequently employed to facilitate a seamless transition between legacy and updated models, thereby maintaining knowledge consistency ~\citep{li2022learning, li2023variational,DBLP:journals/ijcv/XuanYZ25}. 
Parameter isolation strategies ~\citep{schwarz2018progress, zhang2023continual,zhao2023does} offer a different solution by safeguarding existing knowledge through the freezing of parameters associated with past tasks, while allocating new parameters for upcoming tasks. However, this approach can engender increased model complexity and necessitate greater maintenance efforts. 
Rehearsal methods ~\citep{bang2021rainbow, bonicelli2022effectiveness,gao2023dkt} bolster prior learning by retraining on examples from previous tasks, although these methods are often constrained by the need for additional memory and may pose data privacy issues. The ongoing challenges posed by these methods underscore the need for innovative rehearsal-free techniques to effectively alleviate catastrophic forgetting in continual learning systems.

\minisection{Prompt-based Continual Learning.} In contrast to traditional methods that necessitate training a model from scratch, recent advancements in continual learning have shifted towards the fine-tuning of pre-trained networks ~\citep{chen2021pre, pourpanah2022review, liu2023pre, pei2024sa2vp}. This paradigm is increasingly favored for its superior adaptability and learning efficiency. 
Prompt-based fine-tuning has emerged as a prominent strategy within this domain. Approaches such as L2P ~\citep{wang2022learning} and DualPrompt ~\citep{wang2022dualprompt} represent innovative methodologies that leverage learnable prompts to dynamically steer the model’s attention during both training and inference phases. These prompts function as modular instructions that are integrated into the model, thereby encapsulating acquired knowledge and ensuring its ready accessibility, which is crucial for maintaining performance across diverse tasks.
The incorporation of key-query mechanisms, as exemplified by methods like CODAPrompt ~\citep{Smith_2023_CVPR}, further enhances the precision of prompt-based learning by enabling a more nuanced, context-aware application of prompts. This advancement substantially broadens the model’s capacity to manage a wide array of learning scenarios without compromising existing knowledge. Nevertheless, the model’s predictive accuracy can be compromised if irrelevant prompts are selected during the inference process.
To address this issue, CPrompt ~\citep{cprompt} introduces a consistent prompting strategy aimed at enhancing the robustness of prompts, thereby minimizing potential interference and ensuring more reliable performance.

\minisection{Adapter-based Continual Learning.} Adapter architectures ~\citep{houlsby2019parameter, chen2022adaptformer} were initially introduced for the fine-tuning of large language models (LLMs), enabling adaptation to new tasks through the addition of a minimal set of learnable parameters, while preserving the integrity of the backbone network’s parameters. This approach facilitates substantial parameter sharing, thereby enhancing efficiency.
Building upon these foundational strengths, SSIAT ~\citep{tan2024semantically} extends the application of adapters to the realm of continual learning, implementing a prototype shift strategy to counteract catastrophic forgetting induced by adapter parameter updates. This methodology has facilitated efficient fine-tuning of pre-trained networks, yielding performance that surpasses that of prompt-based approaches. C-ADA ~\citep{gao2024beyond} adopts an expandable parameter strategy for adapters, coupled with regularization techniques to steer the trajectory of parameter updates, thereby mitigating interference between newly acquired and previously established knowledge. EASE ~\citep{zhou2024expandable} introduces a novel concept by designing a distinct, lightweight adapter module for each new task, and employs a semantic-guided prototype complement strategy to synthesize new features for existing classes, further contributing to the reduction of catastrophic forgetting.
In contrast to these existing methodologies, our proposed framework innovates by incorporating a semantic prompt to enhance the generalization capacity of image features, and by designing adapters that seamlessly integrate semantic information with visual data. This integration aims to render the features more adaptable to the demands of continual learning tasks, representing a significant advancement in the field.

\begin{figure*}[tb]
\begin{center}
    \includegraphics[width=\textwidth]{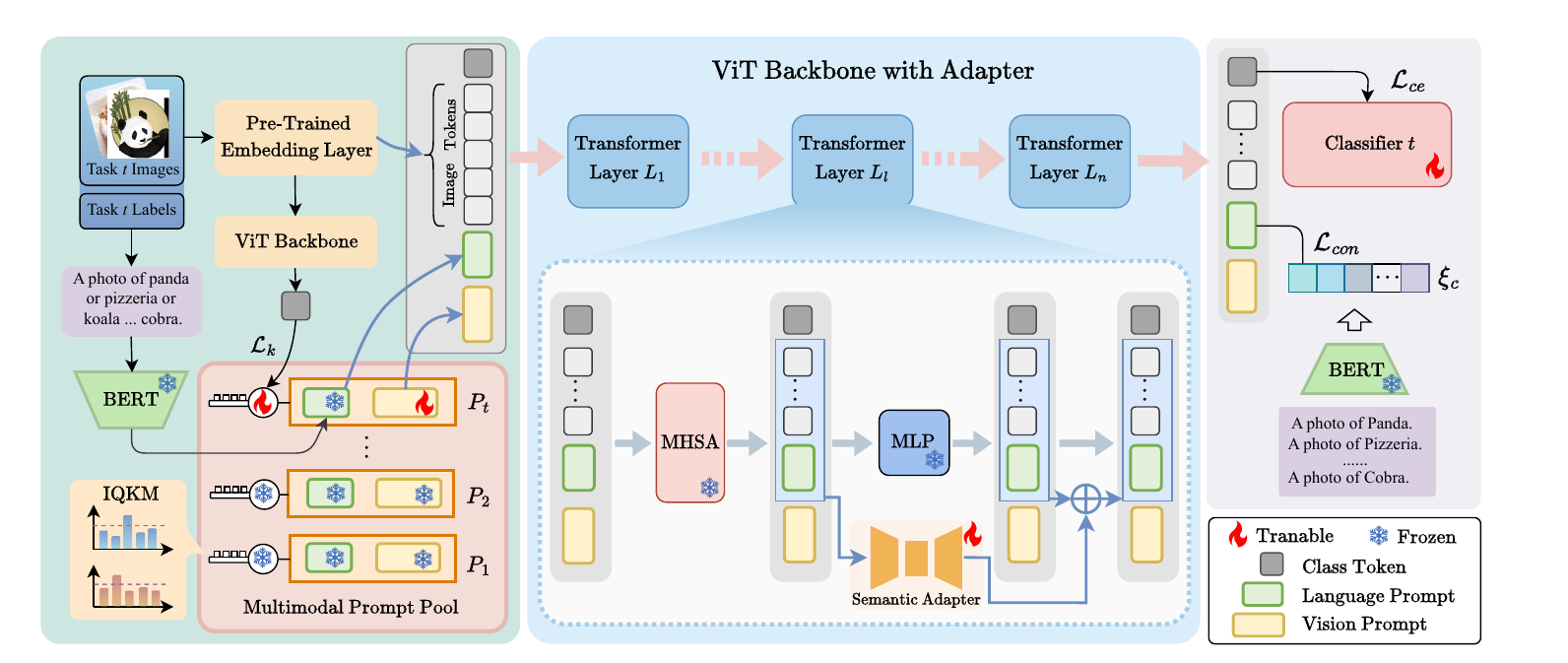}
\end{center}
    \caption{Main architecture of AESP. The semantic prompts generated by a large language model and learnable visual prompts are incorporated to enrich the input information (Section \ref{3.3}). An integrated Query-Key matching mechanism is introduced to select the relevant prompts (Section \ref{3.4}). The selected prompts alongside the image tokens are then fed into a novel backbone network, which includes adapters that can adapt to varying depths of the backbone network for feature refinement and learning (Section \ref{3.5}). The final output features are optimized using classification loss and the proposed cosine contrastive loss to guide the model’s fitting process (Section \ref{3.6}).}
    \label{fig:main}
\end{figure*}

\section{Method}
\label{sec:method}

\subsection{Overview}
We propose a novel Adapter-Enhanced Semantic Prompting framework (AESP). As illustrated in Figure~\ref{fig:main}, the AESP framework comprises three integral modules: the multimodal prompt module, the integrated query-key matching module, and the semantic adapter module. The multimodal prompt module integrates both visual and semantic prompts, leveraging semantic information to augment the generalization capabilities of visual features. The integrated query-key matching module employs a sophisticated combination of matching strategies to enhance the accuracy and reliability of prompt selection. Acknowledging the substantial differences in representation and feature spaces between visual and semantic features, we introduce semantic adapters, strategically embedded within each layer of the Vision Transformer (ViT) network, to facilitate the seamless integration of these diverse information types. Additionally, we propose a novel semantic contrast loss function, designed to enforce the alignment of semantic prompts within a well-defined class-specific semantic space. Each module’s functionality and contribution to the overall framework will be elaborated upon in the following sections.

\subsection{Preliminary}
Continual learning (CL) endeavors to equip models with the capability to assimilate knowledge from sequentially presented tasks with evolving data distributions, while maintaining competence in previously encountered tasks. Consider a dataset $\mathcal{D}$ comprising $N_c$ different classes $\mathcal{D} =\left \{ \mathcal{D}_1, \dots, \mathcal{D}_\mathcal{T} \right \}$, which is partitioned into $\mathcal{T} = N_c / N_{inc}$ incremental tasks, where $N_{inc}$ signifies the number of classes introduced in each successive task phase. The dataset for the $t$-th task is denoted as $\mathcal{D}_t = \left \{ (x_i^t, y_i^t) \right \}_{i=1}^{N_t}$, where $N_t$ is the number of samples, $x_i^t$ is the $i$-th image in the $t$-th task, and the $ y_i^t $ is its corresponding label. The label $y_i \in Y(t)$, where $Y(t)$ represents the class set in task $t$. After the $t$-th session, all the classes learned are denoted as $\mathcal{Y} = Y(1) \cup Y(2) \cup \dots \cup Y(t)$. To emulate real-world data streams, we employ a sequential data-feeding protocol. Our investigation is particularly focused on Class-Incremental Learning (CIL) ~\citep{mittal2021essentials, gao2023dkt, zhou2024class}, a scenario wherein models are trained with explicit task boundaries but are required to perform inference without knowledge of the task identity.
   
We utilize a Vision Transformer (ViT) ~\citep{dosovitskiy2021an}, pre-trained on the ImageNet dataset, as the foundational architecture of our network. This architecture consists of an embedding layer $f_e$ followed by $N_b$ Multi-Head Self-Attention Blocks (MHSA) $\{f_b\}_{b=1}^{N_b}$. For an input image $x \in \mathbb{R}^{H \times W \times C}$ from the dataset $\mathcal{D}_t$, the embedding layer initially converts it into a token sequence, represented as a matrix $\xi_e = f_e(x) \in \mathbb{R}^{L_{img} \times d}$, where $L_{img}$ is the number of tokens and $d$ is the embedding dimension.
Subsequently, this token sequence is augmented with a learnable class token $[CLS] \in \mathbb{R}^{d}$ and through the series of MHSA blocks to yield the final image representation $\xi_{cls} \in \mathbb{R}^{d}$.
This representation $\xi_{cls}$ then used to train task-specific classifiers $\Phi(\cdot; \phi)$, where $\phi \in \left \{ \phi_1, \phi_2, \dots, \phi_{\mathcal{T}} \right \}$ represents the parameters of the classifiers corresponding to each training task.

\subsection{Multimodal Prompt}
\label{3.3}
Distinct from traditional methodologies that solely focus on learning visual prompts for feature adaptation, our approach integrates semantic prompts to augment the generalization capabilities of image features. These semantic prompts are derived from class semantic information. Specifically, we construct a semantic description $R_{lan}$ as follows: "A photo of \{class1\} or \{class2\} ... or \{class$N_{inc}$\}."
For the $t$-th task dataset, $class1$, $class2,  \dots, classN_{inc}$ are substituted with the actual class names from $Y(t)$. We then employ the text encoder of the pre-trained large language model BERT \cite{kenton2019bert} as a semantic feature extractor to transform $R_{lan}$ into a text embedding:
\begin{align}
    P_s = \mathrm{BERT}(R_{lan}),
\end{align}
where $P_s \in \mathbb{R}^{d}$ represents the task-level semantic prompt.

In addition to semantic prompts, we also employ trainable visual prompts to extract classification knowledge from the visual feature space. The visual prompt is defined as $P_v \in \mathbb{R}^{L_{vp}\times d}$, where $L_{vp}$ is the length of the visual prompt.  
$P_v$ is randomly initialized at the outset of training and is subsequently refined during each incremental session, with all trained prompts being stored in a prompt pool for future use.

Then, the input of the ViT backbone network consists of multiple parts:
\begin{align}
    \hat{x}=[\xi_{cls}, \xi_e, P_s, P_v].
\end{align}
where $\xi_{cls}$ is a class token, $\xi_{e}$ refers to the image tokens, $P_s$ and $P_v$ denote the semantic prompt and visual prompt, respectively. 
We define $\hat{x} \in \mathbb{R}^{L_{\hat{x}}\times d}$ to represent the input of the ViT backbone network, and $L_{\hat{x}} = 1+L_{img}+1+L_{vp}$ as the total number of input token lengths.

\subsection{Integrated Query-Key Matching}
\label{3.4}
Traditional prompt-based continual learning frameworks typically involve training task-specific prompts alongside corresponding keys, which are then stored in a shared prompt pool. During inference, an image is first encoded into query representations via a pre-trained Vision Transformer (ViT), and the most relevant prompt is selected by computing similarity scores between the query and learnable key vectors.

However, the significant variability in query features across different classes within the same task complicates the accurate selection of task-specific prompts. To mitigate this issue, recent advancements have introduced a Multi-Key mechanism, which assigns multiple keys to each task. Despite these improvements, precise matching during the initial stages of model training remains a challenge. To further enhance the accuracy of prompt selection, we propose an Integrated Query-Key Matching Mechanism.

Our method is built upon the Multi-Key matching mechanism, where the trainable keys are denoted as:
\begin{align}
    \mathcal{K} = \left \{ \mathcal{K}_1, \mathcal{K}_2, \dots, \mathcal{K}_t, \dots, \mathcal{K}_\mathcal{T} \right \},
\end{align}
where $\mathcal{K}_t = \left \{ K_i^t \right \}_{i=1}^{N_{inc}}$ mean there are $N_{inc}$ keys for each task,and $K_i^t \in \mathbb{R}^{d}$ represents the key of the $i$-th class in $t$-th task.

Initially, the query image $x_i$ is processed by the frozen ViT to extract a query feature $q \in \mathbb{R}^{d}$.
Then, the cosine similarity between query feature $q$ and all keys in $\mathcal{K}_t$ is calculated, resulting in a score vector logit $\hat{\ell}$:
\begin{align}
    \hat{\ell} = [\text{cos}(q, K_1^t), \text{cos}(q, K_2^t), ..., \text{cos}(q, K_{N_{inc}}^t)].
\end{align}

To optimize the alignment of the query feature with its corresponding key while repelling other keys, we employ a cross-entropy loss function \cite{zhang2018generalized} as the multi-key loss:
\begin{align}
    \mathcal{L}_{k} = -\frac{1}{N_{\text{inc}}} \sum_{i=1}^{N_{\text{inc}}} y_i \log\left(\text{softmax}\hat{(\ell_i)}\right),
\end{align}
where $y_i$ is the ground truth label for the input image $x_i$.

Despite this, class disparity-based methods may encounter interference due to similar classes across tasks. How can we effectively discern task-specific characteristics? Experimental observations indicate that for an input $x^t$, the classification logits $\ell^n$ (where $n = t$) exhibit relative stability, typically characterized by a dominant value and several minor values, resulting in low entropy. In contrast, $\ell^n$ (where $n \neq t$) displays a stochastic distribution with higher entropy. This entropy pattern inspires the use of logit entropy as a complementary metric for cross-task matching:
\begin{align}
    H(\ell) = -\sum_{i=1}^{N_{inc}} \ell_{i} \log(\ell_{i}),
\end{align}
where $\ell$ represents the softmax-transformed version of $\hat{\ell}$, with $\ell_{i}$ corresponding to its i-th value. More details are given in Section \ref{4.4}.

Furthermore, we also utilize class prototypes to improve the accuracy of prompt selection. Specifically, we calculate the mean image feature of each class as a prototype $\xi_{q} \in \mathbb{R}^{D}$. For the t-th task, we calculate the similarity between query features and all prototypes, which is subsequently normalized by the softmax function. Consequently, we obtain a new probability distribution vector as:
\begin{align}
    \zeta_{t,c} =\frac{e^{\text{cos}(q, \xi_{q,c})}}{\sum_{j=1}^{N_{inc}} e^{\text{cos}(q, \xi_{q,j})}},
\end{align}
where $c\in\left\{1: N_{inc}\right\}$ represents the index of each class.

During the inference process, the model integrates three strategies to select prompts, where each strategy is formulated as:
\begin{flalign}
    &P_1 = \arg \max_{t \in T} \{\ell_t\}, \\
    &P_2 = \arg \min_{t \in T} \{H(\ell_t)\}, \\
    &P_3 = \arg \max_{t \in T} \{\zeta _t\}, 
\end{flalign}
where $T \in \left\{1:\mathcal{T}\right\}$ represents the number of tasks already trained.
Then, we adopt a voting strategy \cite{breiman2001random} to select the final task prompts. If all outputs differ, the final prompts are chosen by $P_1$.

\subsection{Semantic Adapter}
\label{3.5}

In Section \ref{3.3}, we advocate for the utilization of multimodal prompts to augment the model’s input processing capabilities. However, the representation spaces of semantic and visual information are substantially dissimilar, and the exclusive reliance on the frozen attention modules within the ViT architecture may prove inadequate for the effective integration of multimodal data, thereby potentially limiting the model’s generalization prowess. To alleviate this challenge, we propose the integration of semantic adapters into pre-trained models, facilitating robust visual-semantic interaction and the acquisition of more transferable feature representations. 

A semantic adapter is composed of a down-projection matrix $\mathrm{W}_{down} \in \mathbb{R}^{d \times d'}$, a non-linear activation function such as ReLU, and an up-projection matrix $\mathrm{W}_{up} \in \mathbb{R}^{d' \times d}$. Here, $d'$ refers to the input dimension of the ReLU activation layer. Let $x_{in}$ denote the input to the adapter. A semantic adapter can be formalized as: 
\begin{flalign}
  &x_{out} = \mathrm{W}_{up,L}(\text{ReLU}(\mathrm{W}_{down,L} \cdot x_{in})), \\ 
  &x_{in} = [\xi_e,P_s],
\end{flalign}
where $L$ represents the $L$-th layer of the ViT backbone. By inputting both image tokens $\xi_e$ and the semantic prompt $P_s$ into the semantic adapter, we achieve a synergistic fusion of visual and semantic information. These adapters not only bolster the generalization of image features but also enable precise adaptation of these features to specific tasks.

Building upon the innovative query-key matching mechanism introduced in Section \ref{3.4}, which markedly improves the accuracy of task matching, we employ task-specific adapter parameters to mitigate the forgetting issue that arises from parameter updates. Upon the introduction of a new task, the adapter learns new parameters while preserving the existing parameters for each preceding task. During the inference phase, the query-key matching process is instrumental in selecting both the most pertinent prompts and the corresponding semantic adapters, thereby facilitating efficient and lightweight feature adaptation.

\subsection{Loss Functions}
\label{3.6}

\minisection{Semantic Contrast Loss.} To ensure the semantic consistency and bolster the stability of our models, we introduce a novel Semantic Contrast Loss. For each $i$-th label in ${Y}(t)$ the corresponding class names are systematically structured into class-level semantic expressions, denoted as $R_{sem,i}$. These expressions are subsequently encoded into a semantic space using the BERT model \cite{kenton2019bert}, facilitating a robust representation:
\begin{align}
   R_{sem,i} &=  \text{“A photo of } \left \{class \ i \right\} \text{."}\\
   \xi_{sem,i} &= \mathrm{BERT}(R_{sem,i}).
\end{align}

In light of the substantial heterogeneity present among classes and across domains within the dataset, we advocate for the adoption of a cosine-based function to maintain the consistency of semantic information. The formulation of the proposed Semantic Contrast Loss is expressed as:
\begin{align}
    \mathcal{L}_{\text{con}} = &\frac{1}{N_{\text{inc}}} 
    \sum_{i=1}^{N_{\text{inc}}} \left\{ \hat{y}_i (1-\cos({P}_s, {\xi}_{sem,i}))
    \right. \nonumber \\
    &\left. +\alpha\cdot (1-\hat{y}_i) \left| \cos({P}_s,{\xi}_{sem,i}) \right| \right\},
\end{align}
where 
$\hat{y}_i$ serves as a binary indicator for sample pair classification, with $\hat{y}_i = 1$ denoting a positive sample pair and $\hat{y}_i = 0$ signifying a negative sample pair.

In the context of incremental learning tasks, the prevalence of negative pairs substantially exceeds that of positive pairs, as evidenced by a 10-class incremental phase where the ratio of positive to negative pairs is 1:9. This imbalance presents a challenge for the model to effectively learn from positive samples. To address this issue, we introduce a trade-off factor $\alpha$, which is empirically determined to be 0.3. Through the minimization of the $semantic \ contrast \ loss$, the model is encouraged to produce outputs $P_s$ that are more aligned with their respective class-level semantic features. This alignment guides the adapter within the backbone network to establish a correspondence between semantic information and image representation, thereby enhancing the model’s performance in image classification tasks.

\minisection{Classification Loss.}
Within our framework, each task is equipped with a dedicated trainable classifier $\Phi \left ( \cdot;\phi \right )$, designed to predict the class of an image based on its extracted features $\xi_{cls}$.
Adhering to established practices, we employ the cross-entropy loss function for the optimization of these classifiers, which is mathematically expressed as:
\begin{align}
    \mathcal{L}_{ce} = \text{CrossEntropy}(\Phi \left ( \xi_{cls} ;\phi \right ),y).
\end{align}

\minisection{Final Loss Function.}
To comprehensively capture the learning objectives of our proposed model for a given task $t$, the final loss function is formulated as a composite of three distinct components: the multi-key loss $\mathcal{L}_{k}$, the semantic contrast loss $\mathcal{L}_{con}$, and the cross-entropy loss $\mathcal{L}_{ce}$. This integration ensures a holistic approach to model training, with the overall loss function defined as:
\begin{align}
    \mathcal{L} = \mathcal{L}_{k} + \mathcal{L}_{con} + \mathcal{L}_{ce}.
\end{align}

\begin{table*}[t]
  \centering
  \caption{Performance comparison on Split-ImageNetR under the 10-task setting. 
    The first four rows show prompt-based methods, followed by adapter-based methods, and the proposed AESP in the last row. ${\dagger}$ indicates results reported in the original paper.
    Best results are \textbf{bold}, second-best are \underline{underlined}.}
  \setlength{\extrarowheight}{6pt}
  \begin{tabular}{l@{\hspace{+8mm}}l@{\hspace{+8mm}}l@{\hspace{+8mm}}l@{\hspace{+8mm}}lcc}
    \toprule[1pt]
    \multirow{2}[2]{*}{Method} & \multirow{2}[2]{*}{\shortstack[c]{Reference}} & \multicolumn{3}{c}{Split-ImageNetR} \\
    \cmidrule(lr){3-5}
          &       & Last-acc↑ & Avg-acc↑ & FF↓ \\
    \midrule
    L2P & CVPR 2022 \cite{wang2022learning} & 69.11{\scriptsize ±0.42} & 75.61{\scriptsize ±0.80} & 7.93{\scriptsize±0.05} \\
    DualPrompt & ECCV 2022 \cite{wang2022dualprompt} & 71.50{\scriptsize±0.23} & 76.62{\scriptsize±0.68} & 6.03{\scriptsize±0.88} \\
    CODA-Prompt & CVPR 2023 \cite{Smith_2023_CVPR} & 75.17{\scriptsize±0.23} & 80.59{\scriptsize±0.68} & 7.07{\scriptsize±0.86} \\
    Cprompt & CVPR 2024 \cite{cprompt} & 76.93{\scriptsize±0.55} & 82.45{\scriptsize±0.80} & \underline{5.44}{\scriptsize±0.41} \\
    EASE & CVPR 2024 \cite{zhou2024expandable} & 76.16{\scriptsize±0.25} & 82.85{\scriptsize±0.46} & 7.45{\scriptsize±0.34} \\
    SSIAT$^{\dagger}$ & CVPR 2024 \cite{tan2024semantically} & \underline{79.38}{\scriptsize±0.59} & 83.63{\scriptsize±0.43} & - \\
    ADAM$^{\dagger}$    & IJCV 2025 \cite{DBLP:journals/ijcv/ZhouCYZL25}         &72.87                & 79.39      & - \\
    Lora-DRS$^{\dagger}$ &CVPR 2025 \cite{liu2025lora}      &74.74{\scriptsize±0.78}    & 81.16{\scriptsize±0.59}    & - \\
    DIA$^{\dagger}$ &CVPR 2025 \cite{li2025dynamic}      &79.03    & \underline{85.61}    & - \\
    \textbf{AESP } & This Work & \textbf{82.08}{\scriptsize±0.62} & \textbf{86.08}{\scriptsize±0.52} & \textbf{4.48}{\scriptsize±0.25} \\
    \bottomrule[1pt]
  \end{tabular}%
  \label{main_results}
  \vspace{5mm}
\end{table*}

\begin{table*}[t]
  \centering
  \caption{Comparison of performance on CIFAR100 and Split-ImageNetA under the 10-task setting.}
  \setlength{\extrarowheight}{5pt}
  \begin{tabular}{ll@{\hspace{+6mm}}l@{\hspace{+6mm}}l@{\hspace{+6mm}}l@{\hspace{+6mm}}c@{\hspace{+6mm}}l@{\hspace{+6mm}}l@{\hspace{+6mm}}lll}
    \toprule[1pt]
    \multirow{2}[2]{*}{Method} & & \multicolumn{3}{c}{CIFAR100} &       & \multicolumn{3}{c}{Split-ImageNetA} \\
          &       & Last-acc↑ & Avg-acc↑ & FF↓   &       & Last-acc↑ & Avg-acc↑ & FF↓ \\
\cmidrule{1-1}\cmidrule{3-5}\cmidrule{7-9}
    L2P   & & 82.44{\scriptsize±0.56} & 88.00{\scriptsize±0.97} & 7.04{\scriptsize±1.48} & & 44.29{\scriptsize±0.74} & 53.93{\scriptsize±1.05} & 9.98 {\scriptsize±0.15} \\
    DualPrompt & & 83.07{\scriptsize±0.55} & 88.41{\scriptsize±1.19} & 6.22{\scriptsize±1.43} & & 46.65{\scriptsize±0.19} & 58.06{\scriptsize±0.91} & 12.69{\scriptsize±1.11} \\
    CODA-Prompt & & 86.19{\scriptsize±0.36} & 90.97{\scriptsize±1.19} & 6.70{\scriptsize±0.88} & & 51.90{\scriptsize±0.71} & 62.23{\scriptsize±1.42} & \underline{9.69}{\scriptsize±0.67} \\
    Cprompt & & 87.62{\scriptsize±0.16} & 92.33{\scriptsize±0.28} & \underline{5.20}{\scriptsize±0.32} & & 55.06{\scriptsize±1.19} & 66.25{\scriptsize±1.95} & 12.14{\scriptsize±0.78} \\
    EASE  & & 88.12{\scriptsize±0.56} & 92.49{\scriptsize±0.75} & 5.83{\scriptsize±0.60} & & 54.16{\scriptsize±0.20} & 65.98{\scriptsize±0.78} & 13.05{\scriptsize±0.82} \\
    SSIAT$^{\dagger}$ & & \underline{91.35} {\scriptsize±0.26} & \underline{94.35} {\scriptsize±0.60} & -  & & \underline{62.43} {\scriptsize±1.63} & 70.83 {\scriptsize±1.63} & - \\
    ADAM$^{\dagger}$ & & 85.75    & 90.94 & - & & 49.51 & 59.89 & - \\
    Lora-DRS$^{\dagger}$ & & 89.14{\scriptsize±0.23} & 92.55{\scriptsize±0.25} & - & & - & - & - \\
    DIA$^{\dagger}$ & & 90.8 & 94.29 & - & & 61.69 & \underline{71.58} & - \\
    \textbf{AESP(ours)}  & & \textbf{92.27}{\scriptsize±0.21} & \textbf{95.11}{\scriptsize±0.01} & \textbf{2.95}{\scriptsize±0.27} & & \textbf{64.54}{\scriptsize±0.36} & \textbf{72.16}{\scriptsize±0.43} & \textbf{9.22}{\scriptsize±0.54} \\
    \bottomrule[1pt]
  \end{tabular}%
  \label{main_results1}%
\end{table*}%

\section{Experiments}
    
\subsection{Experimental Settings}
\label{4.1}

\minisection{Datasets and Protocols. }
To facilitate a comprehensive comparison of various continual learning methodologies, we perform a series of experiments across three benchmark datasets, each characterized by distinct challenges, within the context of class-incremental learning ~\citep{mittal2021essentials, cprompt}. In this setting, task identities are obscured during the testing phase, thereby simulating a more realistic scenario. The datasets are delineated as follows:

\begin{itemize}
    \item \textbf{ImageNetR} ~\citep{hendrycks2021many} is a dataset comprising 30,000 images across 200 classes derived from ImageNet. Each class features images in diverse styles, including art, cartoons, graffiti, and particularly challenging examples from the original ImageNet. This variety introduces a significant domain shift, making it especially difficult for models trained on standard datasets.
    \item \textbf{CIFAR-100} ~\citep{krizhevsky2009learning} consists of 60,000 color images at a resolution of 32×32 pixels, categorized into 100 classes, with 500 training images and 100 test images per class. It is a widely recognized benchmark in the continual learning community.
    \item \textbf{ImageNetA} ~\citep{hendrycks2021natural} is a real-world dataset containing 7,500 unmodified, naturally occurring images from 200 ImageNet classes. These images were specifically selected because they are misclassified by ResNet models, posing a significant challenge for machine learning models. Additionally, the dataset exhibits substantial class imbalance, with certain categories containing only a small number of samples.
\end{itemize}

\minisection{Implementation Details.}
Following Cprompt ~\citep{cprompt}, we utilize the Stochastic Gradient Descent (SGD) optimizer ~\citep{keskar2017improving} with a momentum of 0.9 and an initial learning rate of 0.01.
Our experiments were performed on an NVIDIA RTX 3090 GPU, using a batch size of 24 images per iteration. The learning rate is reduced to zero following a cosine annealing schedule.
We trained our model on ImageNetR for 14 epochs and on CIFAR100 and ImageNetA for 20 epochs.

To evaluate our approach in various incremental learning scenarios, we implemented two strategies for ImageNetR and ImageNetA, each comprising 200 classes. The first strategy divides the dataset into 10 tasks, with 20 classes per task, while the second splits it into 20 tasks, each with 10 classes. For CIFAR100, which consists of 100 classes, we divided it into 10 tasks, each containing 10 classes.

\minisection{Evaluation Metrics.}
We assessed our model using three widely adopted metrics in continual learning ~\citep{wang2022learning, wang2022dualprompt, Smith_2023_CVPR, wang2024classsurvey, cprompt}. Specifically, we measure the average prediction accuracy across all classes after the final training session, referred to as Last-acc, and the average accuracy across all sessions, denoted as Avg-acc. We also compute the forgetting score FF, following the approach outlined in ~\citep{wang2022dualprompt, Smith_2023_CVPR}, to evaluate the performance degradation of previous tasks over time. To ensure the reliability and statistical significance of our findings, we conducted each experiment three times, employing different random seeds to vary the dataset order. We report the mean performance metrics of these runs, accompanied by their standard deviations.

\begin{figure*}[tb]
    \begin{center}
        \centerline{\includegraphics[width=\textwidth]{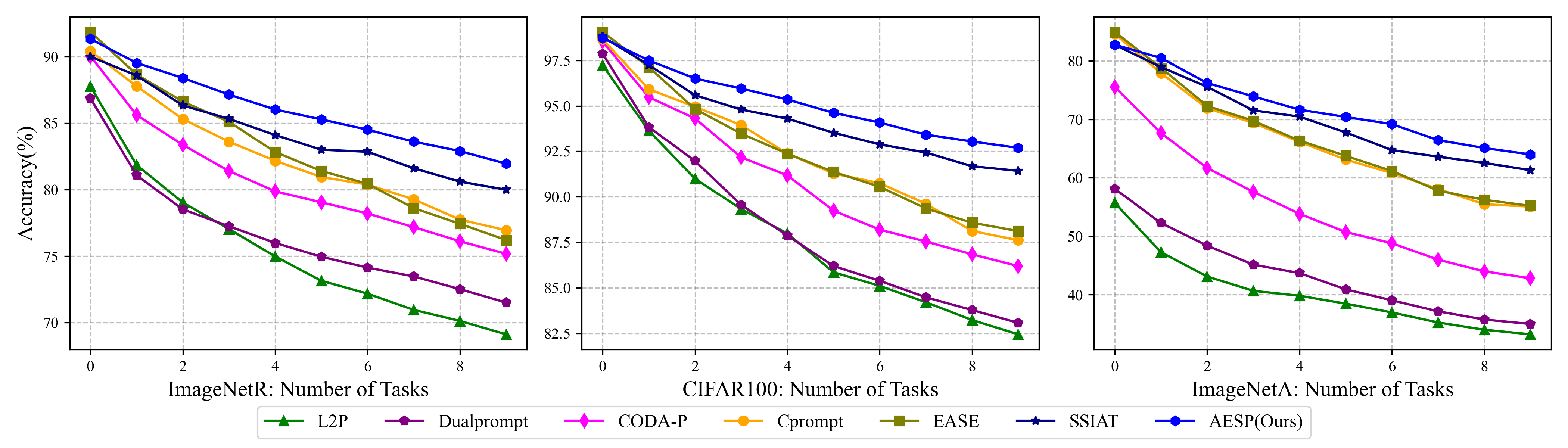}}
    \end{center}
    \vspace{-10mm}
    \caption{Illustrations of the continual learning performance at each task are depicted through these curves. The curves are constructed by averaging the performance across three separate seeds for every incremental learning session.}
    \label{fig:maintable}
\end{figure*}

\begin{table}[t]  
  \centering
     \caption{Experimental results under 20-task setting on Split-ImageNetR and Split-ImageNetA dataset. ${\dagger}$ indicates results reported in the original paper. The best result is marked in \textbf{bold}, and the second-best result is \underline{underlined}.}
\setlength{\extrarowheight}{4pt}
  \setlength{\tabcolsep}{1.5pt}
    \begin{tabular}{llllcll}
    \toprule
    \multirow{2}[2]{*}{Method} & & \multicolumn{2}{c}{Split-ImageNetR} &       & \multicolumn{2}{c}{Split-ImageNetA} \\
          &       & Last-acc↑ & Avg-acc↑ &       & Last-acc↑ & Avg-acc↑ \\
\cmidrule{1-1}\cmidrule{3-4}\cmidrule{6-7}    L2P  & & 65.88{\scriptsize±0.33} & 73.18{\scriptsize±1.37} & & 37.16{\scriptsize±0.36} & 47.95{\scriptsize±0.43} \\
    Dual-P & & 67.80{\scriptsize±0.51} & 73.82{\scriptsize±1.45} & & 39.76{\scriptsize±0.24} & 53.67{\scriptsize±0.72} \\
    CODA-P & & 72.41{\scriptsize±0.37} & 77.98{\scriptsize±1.01} & & 43.78{\scriptsize±0.57} & 56.03{\scriptsize±0.69} \\
    Cprompt & & 74.31{\scriptsize±0.79} & 81.18{\scriptsize±0.15} & & 52.29{\scriptsize±0.81} & 64.02{\scriptsize±1.26} \\
    EASE  & & 74.09{\scriptsize±0.36} & 81.51{\scriptsize±0.47} & & 42.42{\scriptsize±1.29} & 57.36{\scriptsize±1.26} \\
    SSIAT$^{\dagger}$   & & 75.67 {\scriptsize±0.14} & 82.30 {\scriptsize±0.36} & & \underline{59.16} {\scriptsize±1.03} & \underline{68.45} {\scriptsize±1.92}\\
    ADAM$^{\dagger}$  & & 70.47 & 77.29 & & 49.57 & 60.53 \\
    Lora-DRS$^{\dagger}$   & & 74.80 {\scriptsize±0.73} & 80.69 {\scriptsize±0.75} & & - & - \\
    DIA$^{\dagger}$  & & \underline{76.32} & \underline{83.51} & & - & - \\
    \textbf{AESP} & & \textbf{79.83}{\scriptsize ±0.55} & \textbf{84.5}{\scriptsize ±0.56} & & \textbf{61.39}{\scriptsize ±0.49} & \textbf{70.66}{\scriptsize ±0.86}\\
    \bottomrule
    \end{tabular}%
  \label{20task}%
\end{table}%

\subsection{Comparison with state-of-the-arts}
\label{4.2}
 To thoroughly assess the efficacy of our proposed method, we conduct a comparative analysis against leading prompt-based and adapter-based state-of-the-art (SOTA) continual learning techniques, including L2P ~\citep{wang2022learning}, DualPrompt ~\citep{wang2022dualprompt}, CODA-P ~\citep{Smith_2023_CVPR}, CPrompt ~\citep{cprompt}, EASE ~\citep{zhou2024expandable}, and SSIAT ~\citep{tan2024semantically}. These methodologies were chosen due to their representative status at the cutting edge of their respective domains, providing a robust benchmark for evaluation. To ensure a fair comparison, we replicate the optimal results of these methods under identical experimental conditions. The results presented are the average of three independent runs, each with a unique random seed, to ensure statistical reliability. As demonstrated in Table~\ref{main_results} and Table~\ref{main_results1}, the proposed AESP method consistently outperformed state-of-the-art (SOTA) methods across all datasets. Our method shows a substantial improvement over prompt-based approaches. Notably, while Cprompt has achieved significant progress beyond earlier prompt-based techniques, AESP surpasses it on ImageNetR, with a 5.15\% higher Last-acc, a 3.63\% higher Avg-acc, and a 0.96\% lower forgetting rate. Similarly, on CIFAR100, AESP outperforms Cprompt by 4.65\% in Last-acc and 2.78\% in Avg-acc. Moreover, AESP achieves approximately a 2.25\% lower forgetting rate.

Furthermore, although the latest adapter-based methods have outperformed prompt-based approaches, our AESP method, which combines both adapters and prompts, exceeds these adapter-based techniques. Specifically, on ImageNetR, AESP achieves gains of 2.45\% in Last-acc and 2.7\% in Avg-acc over the leading adapter-based method SSIAT. While our advantage in Last-acc over prompt-based methods is notable, our advantage in Avg-acc over adapter-based methods is even more pronounced. This stems from the stronger resistance to the catastrophic forgetting of adapter-based methods. By combining adapters and prompts, AESP effectively reduces the forgetting rate by about 2.97\% on ImageNetR and 2.88\% on CIFAR100 compared to SSIAT.

ImageNetA poses a unique challenge, consisting of samples misclassified by ResNet models due to their adversarial or anomalous nature. However, our method uses semantic information to improve the generalization of image features, improving the classification accuracy of these hard samples. AESP achieves a Last-acc of 64.54\% and an Avg-acc of 72.16\%, reflecting gains of about 2.11\% and 1.33\%, respectively, over SSIAT. A detailed comparison of different continual learning methods for each task is provided in Figure ~\ref{fig:maintable}. It shows that our approach achieves excellent performance on each task and has a lower forgetting rate.
    
Additionally, we assessed AESP under a 20-task setting to evaluate its performance across scenarios involving long sequences of tasks, as shown in Table ~\ref{20task}. Under the 20-task setting, AESP achieves a Last-acc of 79.83\% on ImageNetR and 61.36\% on ImageNetA, sustaining leading performance. These Last-acc results exceed SSIAT by 4.16\% and Cprompt by up to 5.52\%. A similar trend is observed for Avg-acc, indicating that our method maintains nice stability and strong resistance to forgetting when dealing with longer task sequences.

\begin{table}[t]
  \centering
  \caption{Ablation Study on Split-ImageNetR and CIFAR100 dataset. We removed the Semantic Adapter (S-Adapter), Semantic Prompt (S-Prompt), and Integrated query-key matching (IQKM) from the model separately to verify the validity of each component.}
  \setlength{\tabcolsep}{5pt}
  \setlength{\extrarowheight}{5pt}
    \begin{tabular}{lcccc}
    \toprule
    \multirow{2}[2]{*}{Method} & \multicolumn{2}{c}{Split-ImageNetR} & \multicolumn{2}{c}{CIFAR100} \\
\cmidrule{2-5}          & Last-acc↑ & FF↓   & Last-acc↑ & FF↓ \\
    \midrule
    w/o Adapter & 80.45 & \textbf{4.42} & 91.4  & 3.04 \\
    w/o S-Prompt & 81.97 & 4.98  & \textbf{92.37} & 3.06 \\
    w/o IQKM & 74.39 & 8.25  & 84.54 & 6.81 \\
    AESP  & \textbf{82.08} & 4.48  & 92.27 & \textbf{2.95} \\
    \bottomrule
    \end{tabular}%
  \label{ablation}%
\end{table}%

\subsection{Ablation Study}
\label{4.3}
We conducted ablation experiments under the 10-task setting on two datasets to evaluate the effectiveness of three components: semantic adapters, semantic prompts, and integrated query-key matching (IQKM). The results are shown in Table ~\ref{ablation}.
The results show that when the semantic adapters are removed from the ViT layers, the Last-acc decreased by 1.63\% and 0.87\% on ImageNetR and CIFAR100, respectively, while the forgetting score FF remained at a similar level. This indicates that the adapters positively contribute to the model’s recognition accuracy without significantly affecting forgetting.

Moreover, removing the semantic prompts resulted in an increase in FF by 0.5\% on ImageNetR and 0.11\% on CIFAR100, although Last-acc was maintained at a similar level. These findings suggest that semantic prompts assist in feature adaptation, promoting more generalized features that help prevent the forgetting of previously learned knowledge. However, overly generalized features might limit the model’s ability to achieve higher accuracy. The above results collectively demonstrate that AESP successfully achieves a favorable balance between accuracy retention and catastrophic forgetting mitigation through the synergistic integration of semantic adapters and semantic prompts.

Furthermore, when the IQKM module was replaced with the Multi-Key matching mechanism for query-key matching, significant performance drops occurred across both datasets. Specifically, Last-acc decreased by 7.69\% and 7.73\% on ImageNetR and CIFAR100, respectively, and the forgetting score FF increased by 3.77\% and 3.86\%. This substantial decline indicates the importance of IQKM in achieving effective task selection, as detailed in Section ~\ref{4.4}. Inaccurate task selection leads to the use of inappropriate prompts and adapters, resulting in reduced accuracy and increased catastrophic forgetting.

\subsection{Detailed Analysis}
\label{4.4}

\begin{figure*}[t]
    \begin{center}
        \centerline{\includegraphics[width=\textwidth]{figure/x4_four_heatmaps.png}}
    \end{center}
    \vspace{-6mm}
    \caption{Ablation analysis of IQKM for task selection accuracy. Each heatmap represents the correspondence between the query task (ground-truth task) and the selected task. The color intensity indicates the matching accuracy, with darker shades representing higher accuracy.}
    \label{fig:heatmaps}
\end{figure*}

\minisection{Effectiveness of IQKM.}
To enhance prompt selection accuracy, we propose the Integrated Query-Key Matching (IQKM), as detailed in Section~\ref{3.4}. This approach amalgamates Multi-Key matching, Entropy-based matching, and Prototype matching through a consensus voting mechanism to identify the appropriate task. To ascertain the efficacy of IQKM, we conducted an evaluation of task selection accuracy across various methods using the ImageNetR dataset. As shown in Figure~\ref{fig:heatmaps}, we visualize the contribution of each strategy through four heatmaps that reveal the correlation between ground-truth tasks (vertical axis) and model-predicted task selections  (horizontal axis). The diagonal patterns in the heatmaps for three matching approaches indicate substantial task selection errors, particularly evident in their scattered distributions across non-diagonal regions. In contrast, IQKM demonstrates a marked improvement in task selection accuracy, which ensures consistency between knowledge acquisition during training and its application in inference. This alignment between learned representations and task-specific utilization guarantees robust feature encoding stability when handling cross-dataset inputs, preserving task-discriminative patterns that enhance recognition reliability.

\minisection{Detailed Analysis of Semantic Prompt.}
To improve the generalization of image features, we utilize semantic prompts encoded by a text encoder to enrich the input information, as described in ~\ref{3.3}. To validate the effectiveness of semantic embedding, we compare our AESP method with a conventional prompt-based approach, wherein the learnable prompt is randomly initialized and subsequently updated during training. The results, presented in Table ~\ref{tab:sprompt} for the ImageNet-R dataset, indicate that our method achieves a forget rate of 4.48\%, which is significantly lower than that of using trainable prompts. Furthermore, our approach yields a Last-acc of 82.08\% and an Avg-acc of 86.08\%, underscoring the semantic prompt’s efficacy in alleviating catastrophic forgetting and preserving the pre-trained network’s performance. In addition, we introduce a semantic contrastive loss to ensure semantic consistency. To assess its impact, we conducted an experiment replacing the proposed semantic contrast loss with the commonly used cross-entropy loss, with the results displayed in the second line of Table ~\ref{tab:sprompt}. The performance decline observed in the absence of semantic contrastive loss substantiates its critical role in maintaining model performance.

    \begin{table}[t]
      \centering
      \caption{Detail analysis of Semantic Prompt on 10-task continual learning of ImagenetR.}
    \setlength{\tabcolsep}{6pt}
      \setlength{\extrarowheight}{4pt}
        \begin{tabular}{lccc}
        \toprule
        Method & Last-acc↑ & Avg-acc↑ & FF↓ \\
        \midrule
        Trainable & 81.93  & 85.82  & 4.76  \\
        CE Loss & 81.98  & 86.05  & 4.69  \\
        AESP  & \textbf{82.08 } & \textbf{86.08 } & \textbf{4.48 } \\
        \bottomrule
        \end{tabular}%
      \label{tab:sprompt}%
    \end{table}%

\minisection{Influence of text encoder. }
In our prior experiments, we employed the BERT text encoder to transform semantic expressions into a semantic space. To investigate the impact of different text encoders on our AESP method, we further evaluated its performance using two additional pre-trained text transformer models: CLIP ~\citep{radford2021clip} and Sentence-BERT ~\citep{DBLP:Sentence-BERT}. CLIP is designed to learn aligned embeddings for images and text in a shared space, whereas Sentence-BERT creates sentence embeddings optimized for semantic similarity tasks through fine-tuning of BERT. The results for ImageNetR, presented in Table ~\ref{tab:encoder}, reveal that our method exhibits resilience to the choice of text encoder. Notably, CLIP’s text encoder, trained on an extensive corpus of image-text pairs for visual-semantic alignment, achieves the highest Last-Acc score. Sentence-BERT, with its simpler semantic representation, yields a more modest performance improvement. Nonetheless, the consistent performance gains across different text encoders underscore the robustness of AESP to variations in text encoding mechanisms.

\begin{table}[t]
  \centering
  \caption{Influence of different text encoders. We test our proposed method with CLIP, BERT, and Sentence-BERT text encoders on 10-task continual learning of ImagenetR and cifar100.}
  \setlength{\tabcolsep}{6pt}
  \setlength{\extrarowheight}{4pt}
    \begin{tabular}{lcccc}
    \toprule
    \multirow{2}[2]{*}{Encoder} & \multicolumn{2}{c}{ImageNetR} & \multicolumn{2}{c}{CIFAR100} \\
          & Last-acc↑ & FF↓   & Last-acc↑ & FF↓ \\
    \midrule
    CLIP   & \textbf{82.27} & 4.67  & \textbf{92.43} & 3.22 \\
    Sen-BERT  & 82.08 & 4.48  & 92.27 & \textbf{2.95} \\
    Bert  & 81.81 & \textbf{4.47} & 92.4  & 2.97 \\
    \bottomrule
    \end{tabular}%
  \label{tab:encoder}%
\end{table}%

\minisection{Impact of semantic adapter location. }
The semantic adapter plays a crucial role in fusing visual and semantic information, thereby enhancing the generalization capabilities of image features. This adapter can be integrated into various layers of the Vision Transformer (ViT). To determine the optimal number and placement of semantic adapters, we conducted a series of experiments on the ImageNetR dataset within a 10-task setting. The results, as presented in Table ~\ref{layers}, indicate that incorporating adapters into all layers of the ViT yields the highest performance. Furthermore, our findings reveal that placing adapters in shallow layers results in superior outcomes compared to deeper layers. This observation diverges from earlier research ~\citep{gao2024beyond, park2024pre}, which primarily focused on adapter-based methods dealing with pure image information. In our approach, the integration of semantic information suggests that shallow layers may be more conducive to effective feature adaptation, highlighting a distinct advantage of our method in leveraging semantic context.

\begin{table}[t]  
  \centering
    \caption{The influence of Adapters for different layers. Experiments were performed on the Split-imagenetR dataset.}
  \setlength{\tabcolsep}{6 pt}
    \setlength{\extrarowheight}{3 pt}
    \begin{tabular}{lccc}
    \toprule
    Attach Layer & Amount & Last-acc↑ & Avg-acc↑ \\
    \midrule
    Layer \ 0 - 3 & 4     & 81.13 & 85.17 \\
    \rowcolor[rgb]{ .92,  .92,  .92} Layer \ 8 - 11 & 4     & 80.16  & 84.26 \\
    Layer \ 0 - 4 & 5     & 81.28 & 85.17 \\
    \rowcolor[rgb]{ .92,  .92,  .92} Layer \ 7 - 11 & 5     & 81.09 & 84.88 \\
    Layer \ 0 - 5 & 6     & 81.39 & 85.59 \\
    \rowcolor[rgb]{ .92,  .92,  .92} Layer \ 6 - 11 & 6     & 81.24 & 85.44 \\
    \textbf{Layer \ 0 - 11} & \textbf{12} & \textbf{82.08} & \textbf{86.08} \\
    \bottomrule
    \end{tabular}%
  \label{layers}%
\end{table}%

\section{Conclusion}

In this paper, we design an innovative adapter-enhanced semantic prompting framework tailored for continual learning, which synergistically merges the advantages of prompts and adapters to facilitate effective feature adaptation. To enhance the generalization of visual features, we utilize semantic prompts generated by the renowned large language model, BERT, designed to infuse semantic knowledge into the learning process and thereby enrich feature representation. Furthermore, we develop a semantic contrast loss function, essential for preserving semantic consistency across diverse tasks, which aligns the semantic representations of visual features and ensures the model maintains coherent and meaningful interpretations over time.
To optimize the selection of task-specific prompts, we propose an integrated query-key matching mechanism that significantly improves the accuracy of prompt selection—a critical factor for precise image classification in continual learning environments. Our extensive experiments across three established continual learning datasets underscore the efficacy of our proposed framework, validating its potential to advance the field of continual learning.

\section*{Data Availability}

This paper uses public datasets to conduct experiments. Those datasets are available in the ImageNet-R~\url{https://github.com/hendrycks/imagenet-r}, CIFAR-100~\url{https://www.cs.toronto.edu/~kriz/cifar.html}, ImageNet-A~\url{https://github.com/hendrycks/natural-adv-examples}.

\bibliographystyle{spbasic}      %
\bibliography{manuscript}

\end{document}